# k-Nearest Neighbor Optimization via Randomized Hyperstructure Convex Hull

Jasper Kyle Catapang
jcatapang@up.edu.ph
University of the Philippines Manila

*Abstract*—In the k-nearest neighbor algorithm (k-NN), the determination of classes for test instances is usually performed via a majority vote system, which may ignore the similarities among data. In this research, the researcher proposes an approach to fine-tune the selection of neighbors to be passed to the majority vote system through the construction of a random n-dimensional hyperstructure around the test instance by introducing a new threshold parameter. The accuracy of the proposed k-NN algorithm is 85.71%, while the accuracy of the conventional k-NN algorithm is 80.95%, when performed on the Haberman's Cancer Survival dataset—and 94.44% for the proposed k-NN algorithm, compared to the conventional's 88.89% accuracy score on the Seeds dataset. The proposed k-NN algorithm is also on par with the conventional support vector machine algorithm accuracy, even on the Banknote Authentication and Iris datasets, even surpassing the accuracy of support vector machine on the Seeds dataset.

*Index Terms*—k-nearest neighbors, hyperstructure, convex hull

## I. INTRODUCTION

K-NEAREST NEIGHBOR is among the most popular classification and regression algorithm in machine learning. The k-Nearest Neighbor (k-NN) algorithm is among the simplest of all machine learning algorithms as it is easy to implement and only makes use of distance comparisons. Although, the accuracy of k-NN is still relatively low when compared to other classification algorithms. In a particular research [1] support vector machine (SVM) and k-NN were tested against each other and average accuracies of 71.28% and 92.40% for k-NN and SVM respectively were obtained. The relatively low accuracy of k-NN is caused by several factors. One of these factors is that the algorithm has the same result on calculating distance based on the available training instances around the test instances. Another factor that contributes to the low accuracy of the classic k-NN algorithm is the determination of test data classes—which is based on a majority vote system [2], wherein the said system neglects the proximity of data [3]. This poses a significant issue especially when each nearest neighbor are significantly far from the test data [4]. A solution to this problem is to restrict the regions where the test instances can get nearest neighbors from. This research proposes the use of randomly generated hyperstructure boundaries—based on a new threshold parameter—that would ensure the selected nearest neighbors are relatively close to the test instances. The following are the researcher's contributions:

- The researcher presents an approach which limits the selection of neighbors in the conventional k-NN algorithm through the introduction of an n-dimensional bound.
- Subsequently, the approach helps bridge the gap between the accuracies provided by the conventional k-NN algorithm and SVM algorithm, even surpassing both with certain cases [see Table IV and VII].
- The approach offers an optimization without the use of weighted distances for the k-NN algorithm.

In section II, the researcher will be explaining two different approaches that also aim to optimize the accuracy and performance of the k-NN algorithm. In the following section, section III, the nature of the convex hull and the validity of the usage of randomness will be discussed as these deviate from the possible approaches expected. The experimental procedure will then be discussed in detail in section IV, along with the datasets and parameter settings to be used. The results will then be shown in section V, where the different accuracies for the proposed k-NN, conventional k-NN, and conventional SVM on various datasets would be presented. In the succeeding section, the results are to be interpreted and summarized. Next, a conclusion of the research would be shown to concretize the accomplishments of the research. In section VIII, the future work to be done by the researcher is discussed. Lastly, the recommendations are to be shared in section IX.

## II. RELATED WORK

### A. Local Mean Based k-Nearest Neighbor

In [8], local mean based k-nearest neighbor (LMk-NN) is an optimization algorithm proposed. The algorithm proposed is a simple nonparametric classifier. The value of k in LMk-NN is different from the conventional k-NN—conventional k-NN value of k is the number of nearest neighbors from all training data, while in LMk-NN, the value of k is the number of nearest neighbors from each class in the training data [4]. In determining the class of the test dataset, LMk-NN uses the nearest distances to each local mean vector of each data class, which is highly efficient in addressing the negative effects of outliers [4].

## B. Distance Weight k-Nearest Neighbor

In [9], distance weight k-nearest neighbor (DWk-NN) is the other optimization algorithm proposed. Misclassification that may happen because of neglect of the closeness of data can be properly addressed by specifying a new data class based on the weight value obtained from the distance between data. DWk-NN can reduce the influence of outliers and distribution of unbalanced data sets [3].

## III. PRELIMINARY CONCEPTS

In order to better understand the nature of the proposed approach—the use of convex hull and randomization—these concepts are to be discussed for a proper briefing of the procedure to be experimented on the next section. These concepts deviate from the normal optimization approaches for the k-NN algorithm, as it relies on randomness, and therefore need a separate section.

### A. Convex Hull

According to [5], the convex hull of a finite point set $S$ is the set of all convex combinations of its points. In a convex combination, each point $x_i$ in $S$ is assigned a weight or coefficient $\alpha_i$ in such a way that the coefficients are all non-negative and sum to one, and these weights are used to compute a weighted average of the points. For each choice of coefficients, the resulting convex combination is a point in the convex hull, and the whole convex hull can be formed by choosing coefficients in all possible ways. Expressing this as a single formula, the convex hull is the set:

$$\text{Conv}(S) = \left\{ \sum_{i=1}^{|S|} \alpha_i x_i \,\middle|\, (\forall i : \alpha_i \geq 0) \wedge \sum_{i=1}^{|S|} \alpha_i = 1 \right\}.$$

### B. Randomness

Since this research uses a randomized point generator for the hyperstructure, it is important to discuss the randomness behind the generator and why the random number generator used is sufficient for the research's purpose.

To construct a hyperstructure, the researcher has based the simplest structure on two dimensions as a quadrilateral. Its three-dimensional counterpart is a quadrilateral prism, and so on. To make things simple, the base case will be discussed.

The following statements prove that the four points randomly generated are guaranteed to most likely form a quadrilateral. First, a selection of a pair of random points form a line, which can be labeled points $A$ and $B$. The probability that another random point $C$ is virtually zero as the dimension of the line is zero with respect to the plane. This introduces another line. The same can be applied to the last random point $D$—adding two more lines, completing the quadrilateral. To test if at least three points are collinear, the following must be satisfied:

Let the random points $A = (a,b)$, $B = (m,n)$, and $C = (x,y)$. If the line segments AB and BC have the same slope, then A, B, C are necessarily collinear.

The points A, B, and C are collinear if and only if:

$$\det \begin{vmatrix} 1 & a & b \\ 1 & m & n \\ 1 & x & y \end{vmatrix} = 0 \text{ or } (n-b)(x-m) = (y-n)(m-a)$$

These properties hold true for higher dimensions, although the proof is outside the scope of the paper. Additionally, since these properties only work with truly random points, the randomness used by the research also needs to be discussed.

This research uses the Mersenne Twister algorithm, a cryptographically-secured pseudorandom number generator (CSPRNG) developed by Makoto Matsumoto and Takuji Nishimura. A pseudorandom number generator (PRNG) is an algorithm for generating a sequence of numbers whose properties approximate the properties of sequences of random numbers. The PRNG-generated sequence is not truly random, because it is completely determined by an initial value, called the PRNG's seed. Pseudorandom number generators are practically significant because they offer speed in number generation and also reproducibility.[6] CSPRNGs observe special properties that make them cryptographically-secured— these properties are not to be discussed in this paper. The random points generated by the Mersenne Twister algorithm is sufficiently "random" for the four-point argument properties discussed earlier. A problem that illustrates this is called Sylvester's Four-Point Problem. According to [7], Sylvester's Four-Point Problem asks for the probability that four points chosen randomly in a planar region have a convex hull which is a quadrilateral [see Figure 1].

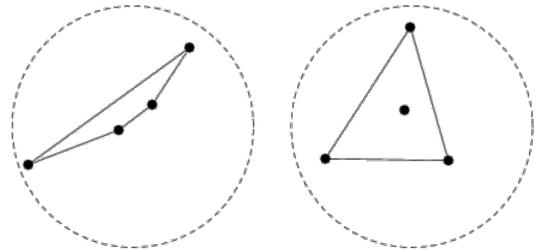

Fig. 1: An illustration of Sylvester's Four-Point Problem

## IV. EXPERIMENTAL STUDY

The following procedure would produce the theorized optimization of k-Nearest Neighbor:

1) Use of values of different k and threshold values
2) Construction of n-dimensional hyperstructure around each test data
   a) Getting minimum value in the n-dimensional coordinate of each test data and subtracting the threshold

- b) Getting maximum value in the n-dimensional coordinate of each test data and adding the threshold
- c) Using the two values from the previous steps as the minimum and maximum for the range to generate random points that will enclose each test data, producing a hyperstructure, similar to a hyperprism, for each test data
- d) Checking if each training data is in their corresponding hyperstructure's convex hull
  - i) If the training data is inside, calculate the Euclidean distance between the training data and the test data
  - ii) If the training data is outside, set its distance to an unreachable value
3) Determination of the containment of each training data inside the constructed n-dimensional hyperstructure
4) Computation of the distance of each test data into each training data
5) For finalizing the value of K and threshold, the training error rate and validation error rate should be computed

This is the pseudocode for the proposed methodology:

```
getNeighbors(threshold, k){
   for each test instance {
      \\let n be the number of dimensions
      distance array = []
      minimum = get minimum among n-
          coordinates of test instance
          point - threshold;
      maximum = get maximum among n-
          coordinates of test instance
          point + threshold;
      generate 4*n-1 random points that
          will form the hyperstructure
      for each training instance {
         if training instance in
             hyperstructure convex hull{
            distance = Euclidean distance
         }
         else distance = unreachable
         save all distances to distance
             array
      }
      sort distance array
      neighbors = get first k elements in
          distance array
   }
}
```

*A. Datasets*

The datasets used are accessible at the Machine Learning Repository of the University of California, Irvine.

The Haberman's Cancer Survival dataset contains cases from a study at the University of Chicago's Billings Hospital on the survival of patients who had undergone surgery for breast cancer.

The Banknote Authentication dataset was extracted from images that were taken from authentic and forged banknotes.

The Iris dataset contains three classes of iris plant of 50 instances each. One class is linearly separable from the other 2; the latter are not linearly separable from each other.

The Seeds dataset is made up of kernels belonging to three different varieties of wheat: Kama, Rosa and Canadian. High quality visualization of the internal kernel structure was detected using a soft X-ray technique.

| Dataset | No. of Instances | No. of Attributes | No. of Classes |
|---|---|---|---|
| Haberman | 306 | 3 | 2 |
| Banknote | 1372 | 5 | 2 |
| Iris | 150 | 4 | 3 |
| Seeds | 210 | 7 | 3 |

TABLE I: Datasets and their respective statistics

The statistics of the different datasets are shown in Table I. These statistics include the number of instances, number of attributes, and number of classes, for each of the dataset used for the experimental study.

*B. Parameter Setting*

The three parameters used for the experimental study are k, threshold, and gamma. The k parameter is to be used by both k-NN algorithms, while the gamma parameter is to be used by the SVM algorithm. The newly introduced threshold parameter would be used by the proposed k-NN algorithm.

The following values would be used for the parameters for each dataset:

| Dataset | k | threshold | gamma |
|---|---|---|---|
| Haberman | 15 | 1.75 | 1e-3 |
| Banknote | 1 | 23 | 1e-3 |
| Iris | 10 | 21 | 0.25 |
| Seeds | 5 | 35 | 0.143 |

TABLE II: Optimal parameter values for each dataset

The following values for the threshold would also be used for another run of the datasets in order to see the effects of wrong selection of threshold value—how it can affect accuracy. The values for k and gamma still hold.

| Dataset | threshold |
|---|---|
| Haberman | 2.5 |
| Banknote | 12 |
| Iris | 15 |
| Seeds | 20 |

TABLE III: Poorly-chosen parameter values for each dataset to demonstrate significant accuracy decrease

## V. EXPERIMENTAL RESULTS

Using different values for k and threshold, the researcher was able to obtain the following results:

| classifier | k | threshold | gamma | accuracy |
|---|---|---|---|---|
| **Proposed k-NN** | 15 | 1.75 | - | **85.71%** |
| Classic k-NN | 15 | - | - | 80.95% |
| **Classic SVM** | - | - | 1e-3 | **85.71%** |

TABLE IV: Accuracies for different parameters and classifiers on the Haberman's Cancer Survival Dataset

| classifier | k | threshold | gamma | accuracy |
|---|---|---|---|---|
| **Proposed k-NN** | 1 | 23 | - | **100%** |
| **Classic k-NN** | 1 | - | - | **100%** |
| **Classic SVM** | - | - | 1e-3 | **100%** |

TABLE V: Accuracies for different parameters and classifiers on the Banknote Authentication Dataset

| classifier | k | threshold | gamma | accuracy |
|---|---|---|---|---|
| **Proposed k-NN** | 10 | 21 | - | **100%** |
| **Classic k-NN** | 10 | - | - | **100%** |
| **Classic SVM** | - | - | 0.25 | **100%** |

TABLE VI: Accuracies for different parameters and classifiers on the Iris Dataset

| classifier | k | threshold | gamma | accuracy |
|---|---|---|---|---|
| **Proposed k-NN** | 5 | 35 | - | **94.44%** |
| Classic k-NN | 5 | - | - | 88.89% |
| Classic SVM | - | - | 0.143 | 83.33% |

TABLE VII: Accuracies for different parameters and classifiers on the Seeds Dataset

| classifier | k | threshold | gamma | accuracy |
|---|---|---|---|---|
| Proposed k-NN | 15 | 2.5 | - | 64.29% |
| Classic k-NN | 15 | - | - | 71.43% |
| **Classic SVM** | - | - | 1e-3 | **78.57%** |

TABLE VIII: Accuracies for different parameters and classifiers on the Haberman's Cancer Survival Dataset with intentional poorly-chosen threshold value

| classifier | k | threshold | gamma | accuracy |
|---|---|---|---|---|
| Proposed k-NN | 1 | 12 | - | 90.90% |
| **Classic k-NN** | 1 | - | - | **100%** |
| **Classic SVM** | - | - | 1e-3 | **100%** |

TABLE IX: Accuracies for different parameters and classifiers on the Banknote Authentication Dataset with intentional poorly-chosen threshold value

| classifier | k | threshold | gamma | accuracy |
|---|---|---|---|---|
| Proposed k-NN | 10 | 15 | - | 81.25% |
| **Classic k-NN** | 10 | - | - | **93.75%** |
| Classic SVM | - | - | 0.25 | 87.5% |

TABLE X: Accuracies for different parameters and classifiers on the Iris Dataset with intentional poorly-chosen threshold value

| classifier | k | threshold | gamma | accuracy |
|---|---|---|---|---|
| Proposed k-NN | 5 | 20 | - | 68.42% |
| **Classic k-NN** | 5 | - | - | **94.74%** |
| Classic SVM | - | - | 0.143 | 84.21% |

TABLE XI: Accuracies for different parameters and classifiers on the Seeds Dataset with intentional poorly-chosen threshold value

## VI. Discussion

The same selection of k for the classic k-NN and proposed k-NN prove to provide the best accuracy in choosing the number of nearest neighbors. The newly introduced threshold parameter used to construct the hyperstructure bounds further refine the accuracy as seen by comparing the accuracy of the proposed classifier on the Haberman's Cancer Survival Dataset—with k=15 and threshold=1.75—and the accuracy of the classic k-NN—with k=15, 85.71% and 80.95% respectively [Table IV]. The Banknote Authentication and Iris datasets results prove that the proposed k-NN holds up even when the classic k-NN and classic SVM have accuracies of 100% [Tables V and VI]. For the Seeds dataset, the proposed k-NN emerges most accurate with a score of 94.44%, compared to classic k-NN: 88.89%, and classic SVM: 83.33% [Table VII].

For the next run of the study, the threshold values are changed to the values in Table III. For the Haberman's Cancer Survival Dataset, the threshold value 1.75 is changed to 2.5. This made the accuracy of the proposed k-NN classifier to be the worst performing with an accuracy of 64.29%. For the Banknote Authentication Dataset, the threshold value 23 is changed to 12. This also made the proposed k-NN classifier's accuracy to be the lowest. The accuracy is 90.90%, compared to the classic k-NN and SVM accuracies of 100%. Next, the proposed k-NN algorithm gave Iris Dataset the worst accuracy of 81.25% compared to classic k-NN: 93.75%, and classic SVM: 87.5%. Lastly, the most drastic drop in accuracy is given by the proposed k-NN algorithm with a threshold value of 20 instead of the optimal value 35. With the optimal value 35, the proposed k-NN algorithm was able to significantly surpass both classic k-NN and SVM, but with the value as 20, the proposed k-NN resulted in an accuracy of 68.42%, compared to classic k-NN: 94.74%, and classic SVM: 84.21%. These results highlight the importance of the proposed methodology's final step: the finalizing of the value of K and threshold using the training error rate and validation error rate by using various values of K and threshold.

## VII. Conclusion

Randomized Hyperstructure Convex Hull k-Nearest Neighbor is an optimized k-Nearest Neighbor algorithm that solves the proximity issues of its predecessor by the use of convex hulls of multiple randomized hyperstructures tailor-fit for each test instance through the introduction of a new threshold parameter. Comparison results demonstrate the availability and effectiveness of the proposed algorithm.

## VIII. Future Work

Randomized Hyperstructure Convex Hull k-NN opens a lot of future work. First, this approach can be combined with either LMk-NN or DWk-NN. The fusion of these optimizations may improve the performance even more. Next, the distance metric used for this approach is only limited to the Euclidean distance metric. Other distance metrics might prove to yield a significant increase in accuracy. The researcher would like to try and integrate these in the future with the never-ending motivation of trying to redeem the k-Nearest Neighbor algorithm with the rest of the machine learning classification algorithms.

## IX. Recommendations

Since this algorithm relies on random points to construct a hyperstructure of which the convex hull would be calculated from, it would be interesting to also try using a fixed structure like an n-dimensional rectangular prism to check if such approach would provide more accuracy than a randomized hyperstructure. Additionally, the use of an n-dimensional sphere could be explored, integrating k-Nearest Neighbor with Radius-Based Nearest Neighbor, and also finding out if this methodology would be an improvement on the algorithm.

## Acknowledgment

The researcher would like to thank his professors, Sir Geoff Solano and Ms. Perl Gasmen, for further motivating him in exploring the world of machine learning, as well as giving him invaluable advice regarding the matter.